\documentclass[letterpaper, 10pt, conference]{ieeeconf}  

\IEEEoverridecommandlockouts                              

\usepackage{graphics} 
\usepackage{epsfig} 
\usepackage{times} 
\usepackage{multicol}
\usepackage[bookmarks=true]{hyperref}

\usepackage{amsmath} 
\usepackage{amssymb}  
\usepackage{bm}  

\usepackage[noadjust]{cite} 
\usepackage{subfigure}
\usepackage{xcolor}
\usepackage{mathtools}
\usepackage{multirow}
\usepackage{booktabs}
\usepackage{upgreek}

\usepackage[capitalise,nameinlink]{cleveref}
\newcommand{\bmat}[1]{\begin{bmatrix}#1\end{bmatrix}}

\def\ie{\emph{i.e.}} 
\def\eg{\emph{e.g.}}

\begin{document}

\title{\LARGE \bf 
User-customizable Shared Control \\ for Robot Teleoperation via Virtual Reality}
\author{Rui Luo$^{1*}$, Mark Zolotas$^{1*}$, Drake Moore$^{1}$, and Ta\c{s}k{\i}n~Pad{\i}r$^{1,2}$
\thanks{*These authors contributed equally to this work.} 
\thanks{This material is based upon work supported by the National Science Foundation under Award No. 1928654.} 
\thanks{$^{1}$Institute for Experiential Robotics (IER), Northeastern University, Boston, Massachusetts, USA. {\tt\footnotesize \{luo.rui, m.zolotas, moore.dr, t.padir\}@northeastern.edu}}%
\thanks{$^{2}$Ta\c{s}k{\i}n Pad{\i}r holds concurrent appointments as a Professor of Electrical and Computer Engineering at Northeastern University and as an Amazon Scholar. This paper describes work performed at Northeastern University and is not associated with Amazon.}}

\maketitle

\begin{abstract}
Shared control can ease and enhance a human operator's ability to teleoperate robots, particularly for intricate tasks demanding fine control over multiple degrees of freedom.
However, the arbitration process dictating how much autonomous assistance to administer in shared control can confuse novice operators and impede their understanding of the robot's behavior. To overcome these adverse side-effects, we propose a novel formulation of shared control that enables operators to tailor the arbitration to their unique capabilities and preferences. Unlike prior approaches to ``customizable'' shared control where users could indirectly modify the latent parameters of the arbitration function by issuing a feedback command, we instead make these parameters \textit{observable} and \textit{directly} editable via a virtual reality (VR) interface. We present our user-customizable shared control method for a teleoperation task in $\mathrm{SE}(3)$, known as the buzz wire game. A user study is conducted with participants teleoperating a robotic arm in VR to complete the game. The experiment spanned two weeks per subject to investigate longitudinal trends. Our findings reveal that users allowed to interactively tune the arbitration parameters across trials generalize well to adaptations in the task, exhibiting improvements in precision and fluency over direct teleoperation and conventional shared control. 
\end{abstract}

\IEEEpeerreviewmaketitle

\section{Introduction}
\label{sec:intro}

Teleoperation is the foundation behind many robotic applications. These applications range from preventing human presence in hazardous environments to providing healthcare services through remote patient care and robot-assisted surgery. Nevertheless, remotely controlling a complex robot warrants a certain level of operator skill and domain expertise in order to successfully complete challenging teleoperation tasks~\cite{darvish2023teleoperation}. Without this baseline, a novice operator may be exposed to detrimental levels of physical and cognitive workload~\cite{Selvaggio2021Autonomy}, potentially resulting in catastrophic task failures~\cite{Luo2023Team}. A common means of relieving this burden on the operator is to employ an autonomous controller that continuously assists the human user by sharing control over the robot~\cite{Losey2018Review}. 

While shared control alleviates excess workload exerted on operators, it may also give rise to new issues that impact task performance and impede dexterity training~\cite{OMalley2005Shared}. One prominent example of such an issue is \textit{model misalignment}. This phenomenon occurs when the division in control between the human and autonomy creates a misunderstanding in how the operator expects the robot to behave~\cite{Zolotas2019XSC}. In direct teleoperation, model misalignment may arise whenever a robot has higher degrees-of-freedom (DoF) than the user's control interface~\cite{Music2017Control}. Despite this interface asymmetry in DoF, a user with adequate training should still gradually build a mental model of the robot's behavior. In contrast, the extra layer of arbitration introduced in shared control exacerbates the user's difficulty in comprehending the system. A communication medium for user feedback is therefore essential to how an operator understands the arbitration~\cite{Losey2018Review}. 

Aside from equipping operators with suitable feedback, another integral aspect of shared control is \textit{user customizability}. Over a long-term interaction, operators will experience variations in their capabilities, preferences, and environments~\cite{Argall2018Autonomy}. Shared control must account for these changes and adapt accordingly, especially in terms of how control authority is assigned~\cite{Li2023Trends}. Prior works have developed promising frameworks for adaptive shared control by enabling users to interactively customize the parameters that characterize the arbitration function~\cite{Gopinath2017Human,Miller2021Analysis} or suggest corrections for the resulting robot behavior~\cite{Hagenow2021Corrective,Cui2023No}. 
However, these works rely entirely on the robot's legible state to convey the effects of a user's modifications to the arbitration procedure. 

\begin{figure}[t]
    \centering
    \includegraphics[width=0.97\columnwidth]{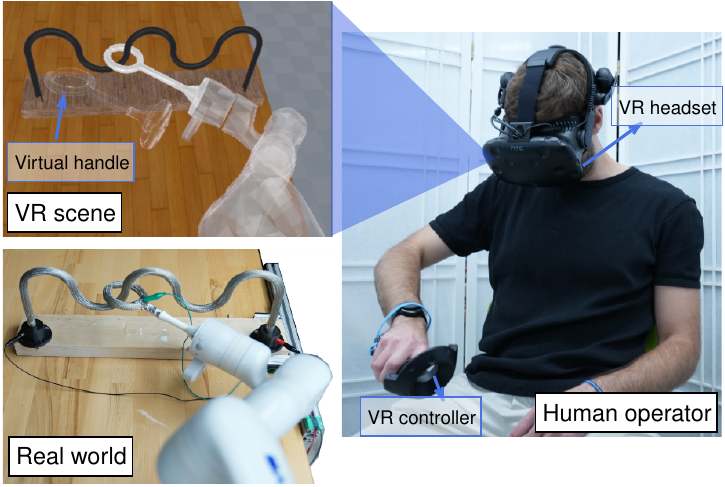}%
    \caption{A user wearing a virtual reality (VR) headset teleoperates the Kinova Gen 3 robotic arm to play the buzz wire game. The VR controller acts as a handle in the virtual world, where its simulated effects also have a direct consequence on the physical counterpart played by the robot.}
    \label{fig:overview}
    \vspace{-3.7mm}
\end{figure}

In this paper, we establish user-customizable shared control by \textit{directly} communicating the internal arbitration parameters to an operator for refinement. This transparency allows users to better understand the arbitration process and thus effectively modify it for personalized outcomes. We ground this idea in a teleoperation task where operators must control a 7-DoF robotic arm in virtual reality (VR) to complete the buzz wire game (see~\cref{fig:overview}). The buzz wire game is a suitable testbed for fine robot control, as it requires both translational and rotational motion, significant eye-hand coordination, and sustained operator focus~\cite{Read2013Binocular,Budini2014Dexterity}. Moreover, the game is frequently utilized for skill training in domains that would benefit from robot teleoperation, such as surgery~\cite{Schreuder2011Surgery}. To assess the quality and longevity of the proposed shared control method, we conducted a user study to analyze how participant performance evolves over repeated interactions. 

The key contributions of this paper are as follows:
\begin{itemize}
    \item A novel mathematical framework that factors in user feedback to formalize user-customizable shared control;
    \item A comprehensive demonstration of this framework within the context of a teleoperation task in 
    ${\mathrm{SE}(3)}$, involving a real 7-DoF robotic arm and a VR interface as the bidirectional communication channel; 
    \item Results from a longitudinal user study with 12 subjects teleoperating the robotic arm to perform the buzz wire game over four sessions spread across two weeks.
\end{itemize}

\section{Related Work}
\label{sec:related}


Sensory feedback can be supplied across multiple modalities in teleoperation. For example, haptic devices are widely used in bilateral teleoperation to simultaneously enhance an operator's task efficiency and encourage growth in skill~\cite{Losey2018Review,Selvaggio2021Autonomy}. Other modalities,~\eg, audio, force, and visual feedback, have enjoyed similar success~\cite{Zeestraten2018Programming,Nicolis2020General,Fitzsimons2020Task}. 
In recent years, VR headsets have proven advantageous in complex manipulation tasks where high-DoF robots are remotely controlled~\cite{Hetrick2020Comparing,Zolotas2021Motion,Babaians2022Skill}. Furthermore, VR interfaces offer operators the opportunity to modify the inner workings of the teleoperated robot, which is an asset we exploit in this work.

In shared control teleoperation, the challenge is to also effectively ``blend'' user inputs with an autonomous controller's outputs. This blending or arbitration process is typically described by parameters that determine how control is allocated. Arbitration parameters may be fixed,~\eg, to favor the human leading, or dynamically updated based on heuristics~\cite{Losey2018Review}, such as safety, fluency, and confidence~\cite{Li2011Dynamic,Dragan2013Blending,Zurek2021Situational}. These parameters can also be personalized to the user by balancing out their capability to independently complete a task with their need for assistance~\cite{Pehlivan2016Minimal,Jain2019Probabilistic}. However, striking the desired balance in autonomy is an open problem~\cite{Li2023Trends}, as heuristically tuning arbitration parameters may not be suitable for all users -- even when the tuning is personalized.


Therefore, it may be more appropriate to empower users with the ability to \textit{customize} the arbitration process. While the literature on customizable shared control is sparse, some previous works have successfully granted this ability via user feedback~\cite{Gopinath2017Human,Miller2021Analysis} or ``corrections''~\cite{Hagenow2021Corrective,Cui2023No}. Hence, users can haptically or verbally correct arbitration parameters on a trial-by-trial basis,~\eg, requesting ``less assistance'' for the next interaction. Although effective, these works relied solely on the robot's appearance and movement to communicate the underlying autonomous assistance, which is often insufficient at explaining arbitration to users~\cite{Jain2019Probabilistic,Brooks2020Vis}. Instead, our paper opts for greater transparency by externalizing the factors that influence arbitration and making them editable to operators.

\section{User-customizable Shared Control}
\label{sec:user}

In the following, we propose a mathematical framework for shared control that enables a user to dictate how the robot autonomy should provide teleoperation assistance. 

\subsection{Problem Setup}
\label{sec:user:form}

The shared control system considered in this work is driven by the following interaction data. At time $t$, a human operator's measured state, $\mathbf{x}_h(t) \in \mathbb{R}^{n_h}$ (\eg, joystick deflections), is fed into a control interface to produce control commands, $\mathbf{u}_h(t) \in \mathbb{R}^{m_h}$, for direct robot teleoperation. Similarly, a robot's state and the autonomy's assistive commands are $\mathbf{x}_r(t) \in \mathbb{R}^{n_r}$ (\eg, joint angles) and $\mathbf{u}_r(t) \in \mathbb{R}^{m_r}$, respectively. 
The output commands of the shared control system, $\mathbf{u}_{sc}(t) = \bm{\beta}_{\bm{\theta}}(\mathbf{u}_r(t), \mathbf{u}_h(t))$, adhere to an arbitration function, $\bm{\beta}(\cdot, \cdot)$, parameterized by $\bm{\theta} \in \mathbb{R}^L$. For instance, a linear blending scheme can be written as:
\begin{equation} \label{eq:sc_lb_arb}
\bm{\beta}_{\bm{\theta}}(\mathbf{u}_r(t), \mathbf{u}_h(t)) =  (1 - \alpha_{\bm{\theta}})\mathbf{u}_r(t) + \alpha_{\bm{\theta}} \mathbf{u}_h(t),
\end{equation} 
where $\alpha_{\bm{\theta}} \in [0,1]$ is the blending variable~\cite{Gopinath2017Human}. While $\alpha$ is usually a scalar, we will adopt a matrix representation for more granular authority over the system behavior. Let $\mathbf{A}_{\bm{\theta}} = \text{diag}(\alpha_1, \ldots, \alpha_{m_r})$ be a positive definite diagonal arbitration matrix, with $\forall i, \alpha_{i} \in [0, 1]$,  such that linear blending is:
\begin{equation} \label{eq:sc_arb_multi}
    \bm{\beta}_{\bm{\theta}}(\mathbf{u}_r(t), \mathbf{u}_h(t)) = (\mathbf{I} - \mathbf{A}_{\bm{\theta}})\mathbf{u}_r(t) + \mathbf{A}_{\bm{\theta}} \mathbf{u}_h(t),
\end{equation}
where $\mathbf{I}$ is an identity matrix of appropriate dimensions.


Selecting the parameters, $\bm{\theta}$, of an arbitration function, $\bm{\beta}_{\bm{\theta}}(\cdot, \cdot)$, is critical in shaping the robot's assistive behavior. Instead of manually selecting a fixed $\bm{\theta}$, an ``optimal'' selection is often derived by solving an optimization problem with a sensible objective, $\Gamma$, like minimizing human effort. The issue with this approach is that $\Gamma$ may not be the true cost function of the shared control, $\Gamma^*$, nor would a tractable solution necessarily exist~\cite{Gopinath2017Human}.

\subsection{Feedback-informed User Optimization}
\label{sec:user:inter}


To circumvent the intractable problem of identifying $\Gamma^*$ and its ``optimal'' parameters, we present a method that allows users to customize the arbitration function by configuring their preferred $\bm{\theta}$. The core concept of our method is to establish a \textit{bidirectional} communication channel between the human and robot. We formalize this channel as a vector-valued function $\bm{\psi}(\cdot, \cdot)$ (dropping $t$ for brevity):
\begin{equation} \label{eq:psi}
\bm\theta \leftarrow \bm{\psi}(\mathbf{v}_r, {v}_h),
\end{equation}
where the robot administers feedback, $\mathbf{v}_r$, to the operator about the shared control, while the user performs parameter updates via an interface action, $v_h$. For example, parameter updates could be obtained from discrete verbal commands made by the operator, $v_h \in \mathbb{Z}^+$, \eg, ``more/less assistance''. Most previous efforts at user customization of $\bm \theta$ have solely depended on the robot's legible state, $\mathbf{v}_r = \mathbf{x}_r$, for operator feedback~\cite{Gopinath2017Human,Miller2021Analysis}.

\begin{figure}[t]
    \centering
    \includegraphics[width=0.89\columnwidth]{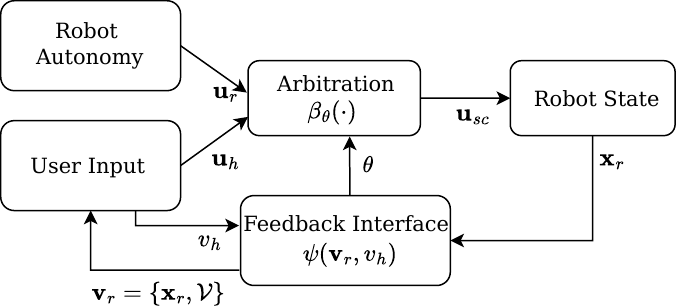}
    \caption{The user-customizable shared control architecture.}
    \label{fig:architecture}
    \vspace{-3.5mm}
\end{figure}

Using this channel formulation to obtain parameters, $\bm \theta$, we represent the arbitration matrix, $\mathbf{A}_{\bm{\theta}}$, as follows:
\begin{align} \label{eq:psi_alpha}
\mathbf{A}_{\bm{\theta}} &= \operatorname{diag}(\bm \alpha) = \operatorname{diag}(\alpha_1, \ldots, \alpha_{m_r}), \\
\text{where} \quad \alpha_j &=  \sum_{l=1}^L w_l \psi_l(\mathbf{v}_r,v_h), \quad j=1, \ldots, m_r \nonumber
\end{align}
Each scalar blending quantity, $\alpha_j$, is governed by potentially more than one parameter, $\theta_l$, output as an element of the channel, $\psi_l(\mathbf{v}_r,v_h)$, depending on the selection weight, $w_l$. Our overall framework architecture is shown in~\cref{fig:architecture}.

Since the feedback variables, $\mathbf{v}_r$ and $v_h$, dramatically impact how users perceive the shared control, we recommend two requirements for generating these variables. First, the robot should communicate more than its external state, $\mathbf{x}_r$, to help operators tune $\bm\theta$. In~\cref{sec:instant}, we demonstrate how a set of human-interpretable cues, $\mathcal{V}$ (\ie, visualizations in VR), can be incorporated into $\mathbf{v}_r=\{\mathbf{x}_r, \mathcal{V}\}$. Second, an operator's parameter updates should have a proportionate effect on robot behavior. This proportionality can be achieved by letting the operator submit actions, $v_h \in [0,1]$ (\eg, using a graphical slider in VR), over the same constrained range as the arbitration parameters, $\forall l, \theta_{l} \in [0, 1]$. 


\section{Proposed Framework Instantiation}
\label{sec:instant}

To illustrate the operational principles of the proposed framework, this section provides an instantiation for assistive teleoperation in the scope of the buzz wire game.

\subsection{Testbed: Buzz Wire Game}
\label{sec:instant:testbed}

In the buzz wire game, the operator's goal is to carefully guide a loop handle in $\mathrm{SE}(3)$ from start to end while minimizing collisions with the wire. The game is a well-known assessment environment for fine motor skill and eye-hand coordination~\cite{Read2013Binocular,Budini2014Dexterity}. Moreover, 3D buzz wire games are advantageous when evaluating dexterity tasks that benefit from stereoscopic depth perception~\cite{Read2013Binocular}. 
As precise motor control, eye-hand coordination, and 3D spatial awareness are critical capabilities for teleoperation in general, we deem this game to be a pertinent testbed. Immersive VR versions of the game have also been validated as an effective simulation for operator training,~\eg, in stroke rehabilitation~\cite{Christou2018Virtual}. 
 

\cref{fig:overview} demonstrates our buzz wire game replica for robot teleoperation. The layout is comprised of a Kinova Gen 3 arm with a loop handle extension at its end-effector to complete the physical game, as well as a person wearing an HTC Vive headset to perform the same task in VR with a handheld controller. The virtual scene is an environment model reconstruction of the buzz wire game and robot, which has been shown to be less cognitively demanding and more usable in VR contexts than point cloud or image renderings~\cite{Wonsick2021Telemanipulation}. An electric circuit board on the physical setup also imitates the game's ``buzz'' feature by detecting collisions. Collisions are then relayed in VR via flashing red lights and vibrations in the handheld input device. 

\subsection{Shared Assistance in Teleoperation}
\label{sec:instant:shared}

In teleoperation, a common strategy is to control the robot's end-effector pose, $\mathbf{x}_r\in\mathbb{R}^6$, by sending desired twist commands, $\dot{\mathbf{x}}_d\in\mathbb{R}^6$, via admittance control:
\begin{equation}
\label{eq:admittance_control}    
    \dot{\mathbf{x}}_{d} = \mathbf{M}^{-1}\int(\mathbf{k}\odot\mathbf{e} + \mathbf{d}\odot\dot{\mathbf{e}} + \mathbf{f}_\mathrm{ext})\, \mathrm{d}t,
\end{equation}
where $\mathbf{M}\in\mathbb{R}^{6\times6}$, $\mathbf{k}\in\mathbb{R}^6$, and $\mathbf{d}\in\mathbb{R}^6$ are mass, stiffness, and damping terms, respectively, while $\mathbf{f}_\mathrm{ext} \in \mathbb{R}^6$ is the externally applied force, $\mathbf{e}\in\mathbb{R}^6$ is the pose error vector, and $\odot$ denotes component-wise multiplication.
In our buzz wire setup, $\mathbf{e}$ is the difference between the handheld VR controller's pose, $\mathbf{x}_h \in \mathbb{R}^6$, and the robot's state, 
${\mathbf{x}_r\in\mathbb{R}^6}$. 

Robot autonomy can then be integrated into~\cref{eq:admittance_control} as an assistive force by reformulating the linear blending in~\cref{eq:sc_arb_multi} to create an admittance control policy for shared control:
\begin{align} \label{eq:v_sc}
    \mathbf{u}_{sc} = \dot{\mathbf{x}}_d &= (\mathbf{I} - \mathbf{A}_{\bm{\theta}})\mathbf{u}_r + \mathbf{A}_{\bm{\theta}} \mathbf{u}_h, \\
    \mathbf{u}_h & = \mathbf{M}^{-1}\int(\mathbf{k}\odot\mathbf{e} + \mathbf{d}\odot\dot{\mathbf{e}})\,\mathrm{d}t, \label{eq:v_sc:h}\\
    \mathbf{u}_r & = \mathbf{M}^{-1}\int \mathbf{w}_{a}\, \mathrm{d}t.
    \label{eq:v_sc:fa}
\end{align}
Here, $\mathbf{f}_\mathrm{ext}$ is replaced by a wrench, $\mathbf{w}_{a}\in\mathbb{R}^6$, that represents autonomous assistance in translational and rotational space.

\subsection{Real-time Assistive Wrench}
\label{sec:instant:real}

Generating real-time autonomous assistance for teleoperation is vital in ensuring reactive and smooth robot motion. To fulfil this requirement, we efficiently compute the assistive wrench, $\mathbf{w}_{a}$, using potential fields, where targets produce attractive vectors and obstacles yield repulsive vectors. Attractive and repulsive vectors are typically calculated between a pair of control and environment points~\cite{flacco2012depth}. In our setting, we define \textit{eight} control points as the vertices of an octagon that encloses the circular end-effector. The environment points are acquired in real-time given point cloud data detected using two depth cameras.

Environment points within a pre-defined range of the control points compose a neighborhood set, $\mathcal{C}=\mathcal{C}_\mathrm{att} \sqcup \mathcal{C}_\mathrm{rep}$, where $|\mathcal{C}|=N$. The $\mathcal{C}_\mathrm{att}$ subset contains attractive points, whereas $\mathcal{C}_\mathrm{rep}$ contains repulsive points.
The potential forces between an environment point, $\mathbf{p}_k\in{\mathcal{C}}$, and a control point, $\mathbf{p}_i\in\mathbb{R}^3$, are computed based on a logistic function:
\begin{equation}\label{eq:f_rep}
    \mathbf{f}_{i,k} =\frac{f_{\mathrm{max}}}{1+\exp\left[ -\lambda _{i,k}\left(\frac{\Vert \mathbf{d}\Vert }{\rho } -d_{i,k}\right)\right]} \ \hat{\mathbf{d}}_{i,k},
\end{equation}
where each function-shaping parameter ${\lambda _{i,k} \in \{\lambda _{\mathrm{rep}} ,\lambda _{\mathrm{att}}\}}$, ${d_{i,k} \in \{d_{\mathrm{rep}} ,d_{\mathrm{att}}\}}$, and ${\hat{\mathbf{d}}_{i,k} =\pm \frac{\mathbf{d}_{i,k}}{\Vert \mathbf{d}_{i,k}\Vert }}$ has two possible values depending on whether the relationship between $\mathbf{p}_k$ and $\mathbf{p}_i$ is repulsive or attractive. Hence, forces $\mathbf{f}_{i,k}$ are either repulsive or attractive, with $f_\mathrm{max}$ determining the maximum magnitude of the generated force and $\rho$ adjusting the change rate in force given distance.

\begin{figure}[t]
    \centering
    \subfigure[Attractive force $\mathbf{f}_{i,k}$ from attractive points on the wire.]{
        \includegraphics[width=0.43\columnwidth]{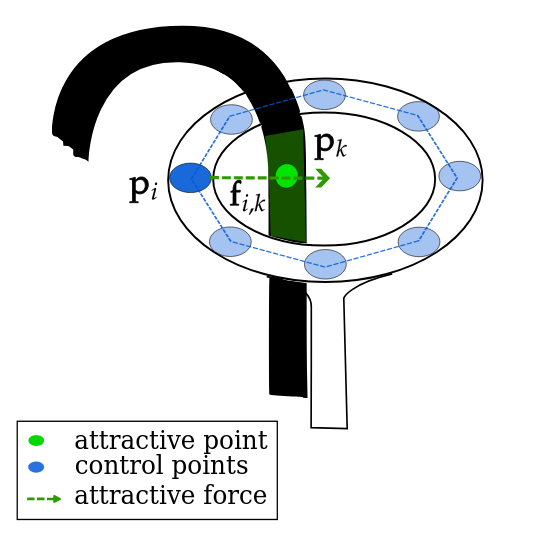}
        \label{fig:assist_wrench:att}
    }
    \subfigure[Repulsive torque $\bm{\tau}_{i,k}$ from sample collision points on the wire.]{
        \includegraphics[width=0.43\columnwidth]{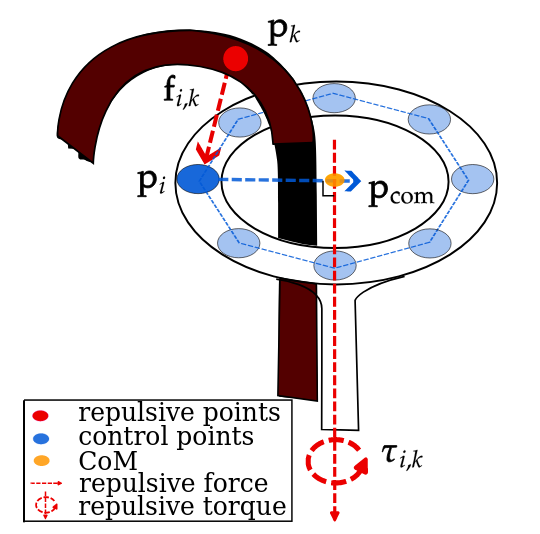}
        \label{fig:assist_wrench:rep}
    }
    \caption{Illustration of how the attractive and repulsive forces are computed given a control point $\mathbf{p}_i$ (dark blue) and environment point $\mathbf{p}_k$ (green/red). 
    }
    \label{fig:assist_wrench}
    \vspace{-4.8mm}
\end{figure}

As the buzz wire game requires 6-DoF teleoperation, we compute the assistive torque $\bm{\tau}_{i,k}$ for rotational motion as:
\begin{equation}
    \bm{\tau}_{i,k} = \mathbf{f}_{i,k} \times (\mathbf{p}_\mathrm{com} - \mathbf{p}_i),
 \end{equation}
where $\mathbf{p}_\mathrm{com}$ is the center of mass of the robot's end-effector. \cref{fig:assist_wrench} illustrates an example of how attractive forces and repulsive torques are calculated for the buzz wire game from a pair of control and environment points $(\mathbf{p}_i,\mathbf{p}_k)$.

The resultant assistive wrench, $\mathbf{w}_{a} = \bmat{\mathbf{f}_\mathrm{net}^\mathsf{T} & \bm{\tau}_\mathrm{net}^\mathsf{T}}^\mathsf{T}$, is derived by aggregating a force component, ${\mathbf{f}_\mathrm{net}\in\mathbb{R}^3}$, and a torque component, $\bm{\tau}_\mathrm{net}\in\mathbb{R}^3$, each calculated as:
\begin{align}
    \mathbf{f}_\mathrm{net} &= \sum_i\frac{\mathbf{f}_i}{\|\mathbf{f}_i\|}\cdot \max_k{\|\mathbf{f}_{i,k}\|}, \\
    \bm{\tau}_\mathrm{net} &= \sum_i\frac{\boldsymbol{\tau}_i}{\|\boldsymbol{\tau}_i\|}\cdot \max_k {\|\boldsymbol{\tau}_{i, k}\|},
\end{align}
where ${\mathbf{f}_i = \sum_k \mathbf{f}_{i,k}}$ and ${\boldsymbol{\tau}_i = \sum_k \boldsymbol{\tau}_{i,k}}$.

\subsection{User-customizable Arbitration}
\label{sec:user:interface}
\begin{figure}[t]
    \centering
    \includegraphics[width=0.68\columnwidth]{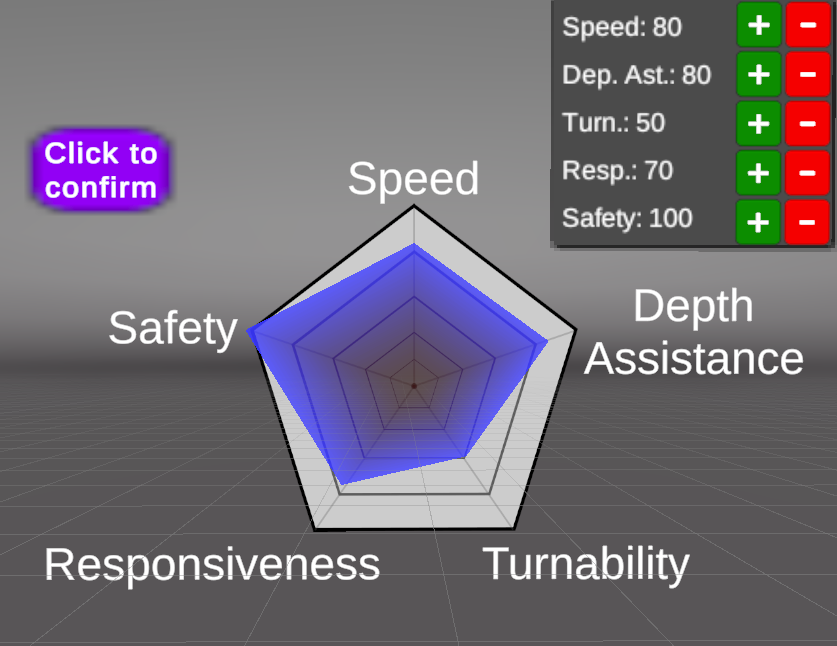}
    \caption{The customizable spider chart interface in virtual reality. The operator can modify any of the five factors in steps of $5$ before confirming their choice. Each factor is scaled from $[0.1, 1]$ to $[10, 100]$.}
    \label{fig:spider_chart}
    \vspace{-4.7mm}
\end{figure}

directional communication, $\bm{\psi}(\mathbf{v}_r,v_h)$. We created a spider chart interface, depicted in~\cref{fig:spider_chart}, to augment the robot's feedback, $\mathbf{v}_r$, with a variety of human-readable indicators, $\mathcal{V}$. Before introducing these indicators, we emphasize that the spider chart is \textit{only visible} when a user wishes to edit the shared control. Hence, no extra cognitive workload is induced during a game trial.

The $\mathcal{V}$ set consists of three arbitration factors displayed in VR: \textit{speed}, \textit{depth assistance}, and \textit{turnability}. \textit{Speed} regulates translational velocity along axes of motion relevant to the buzz wire game (\eg, horizontally along the wire). \textit{Depth assistance} governs the margin between the end-effector and wire along the depth axis, with the aim of assisting line-of-sight occlusions. \textit{Turnability} adjusts the rotational speed of the end-effector. Users can therefore examine the current parameters, $\bm\theta$, via the spider chart, as well as supply edits per factor, $v_h$, in increments/decrements (+/- buttons in~\cref{fig:spider_chart}). Although we defined these three factors for the buzz wire game, regulating motion in relevant DoF is widely used in shared control~\cite{Li2023Trends} (\eg, virtual fixture methods), and maintaining a clear line-of-sight is crucial for robot teleoperation in general~\cite{Nicolis2020General}. We thus believe these arbitration factors can be easily adapted to many different robot teleoperation tasks.

To apply user edits $v_h$ to the arbitration, we express $\bm \alpha \in \mathbb{R}^6$ from~\cref{eq:psi_alpha} as a linear combination:
\begin{align} 
    \label{eq:arbitration-mask}
    \bm \alpha  = \mathbf{W} \bm \theta &= \mathbf{W} \bm{\psi}(\mathbf{v}_r,v_h) = \mathbf{W}\bmat{v_h^\mathrm{speed} \\ 1 - v_h^\mathrm{depth} \\ v_h^\mathrm{turn}}, \\
    \text{where} \quad \mathbf{W} &= \bmat{1 & 0 & 1 & 0 & 0.4 & 0.2\\
          0 & 1 & 0 & 0.2 & 0 & 0\\
          0 & 0 & 0 & 0.8 & 0.6 & 0.8}^\mathsf{T}. \nonumber
\end{align}
The coefficient matrix, $\mathbf{W} \in \mathbb{R}^{6\times 3}$, is structured in a manner where each editable facet, $v_h$, directly maps to a parameter with a semantically similar effect on arbitration. The weighting of elements was chosen empirically to fit the task. For example, $v_h^\mathrm{turn}$ is masked by $\mathbf{W}$ to contribute to an $80\%$ change in arbitration on rotation about the z-axis (yaw), while $v_h^\mathrm{speed}$ contributes to the remaining $20\%$. Here, the term \textit{turnability} intuitively represents more influence over robot rotation than translational speed. The weighting criteria for \textit{depth assistance} uses the complement of $v_h^\mathrm{depth}$, as ``assistance'' implies more contribution from the autonomy.

We also expose two task-independent factors through $\mathcal{V}$: \textit{safety} and \textit{responsiveness}. These factors adjust $f_\mathrm{max}$ in~\cref{eq:f_rep} and $\mathbf{k}$ in~\cref{eq:v_sc:h}, respectively, akin to a variable impedance controller. Using \textit{safety} to tune the force magnitude and \textit{responsiveness} to adjust the stiffness are generalizable to any admittance control scheme. Each of the five configurable factors in $\mathcal{V}$ is constrained in the range $[0.1,1]$ to prevent a ``full autonomy'' configuration. 

\section{Experiment}
\label{sec:exp}

To gauge the effectiveness of the proposed framework, we conducted a between-subjects study involving 12 subjects (3 female; aged 20-34). Subjects were asked to complete the buzz wire game by teleoperating a robotic arm in VR. All participants provided written consent prior to data collection.

\subsection{Experiment Protocol}
\label{sec:exp:prot}

Participants were allocated into one of three groups, each designed to evaluate a different control strategy: (1) direct teleoperation (\textbf{teleop}), which is implemented by setting $\mathbf{A}_{\bm\theta} \, {=} \, \mathbf{I}$ in~\cref{eq:v_sc}; (2) shared control with a heuristics-based arbitration function (\textbf{sc}), as outlined in~\cref{sec:exp:sc}; and (3) our user-customizable shared control method (\textbf{sc\_user}). Participants were unaware of the distinctions between each group. Subjects in \textbf{sc} and \textbf{sc\_user} were provided with identical written explanations of the five factors from~\cref{sec:user:interface} before the experiment. For instance, ``Increases/Decreases in this parameter mean more/less robot assistance in avoiding obstacles'' explained \textit{safety}. It is also worth noting that an equal distribution of the subject pool reported no background in robotics, which was preserved across control strategies. 

\begin{figure}
    \centering
    \includegraphics[width=0.78\columnwidth]{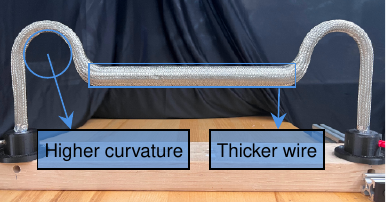}
    \caption{New wire testbed for the last session to assess skill generalization.}
    \label{fig:thick-wire}
    \vspace{-1.7mm}
\end{figure}

Each group consisted of four participants and every subject carried out five consecutive game trials in a single session using their assigned control strategy. This procedure was repeated for a total of four sessions. Subjects were also offered five minutes before their first session to become familiar with the VR setup while not playing the game. This included a brief overview of the spider chart shown in~\cref{fig:spider_chart} for the \textbf{sc} and \textbf{sc\_user} groups. To evaluate the learning rates of operators over an extended period, a minimum of one workday separated each session, resulting in an experiment spanning two weeks. Additionally, the last session involved a distinct wire testbed, as portrayed in~\cref{fig:thick-wire}. This alteration aimed to assess whether any teleoperation skills gained are transferable to a related, yet different task. Subjects also answered post-session surveys on their user experience.

After each trial, participants could spend a minute operating the arm without playing the game. For the \textbf{teleop} group, this post-trial time could be used as an opportunity to improve their understanding of direct robot teleoperation. Participants in \textbf{sc} could instead examine the automatically updated arbitration factors (see~\cref{sec:exp:sc}) through the spider chart and test out the resulting control. Only the \textbf{sc\_user} group could use this post-trial time to edit the arbitration factors displayed on the spider chart and explore different configurations. User-selected arbitration settings persisted across trials unless altered. During game trials, the spider chart was \textit{not} visible and arbitration settings were fixed. 

\subsection{Heuristics-based Shared Control}
\label{sec:exp:sc}

We implemented a heuristics-based assist-as-needed controller as the reference method, $\textbf{sc}$, for our experiment.
The arbitration parameters, $\bm\theta$, in this controller are iteratively updated according to the operator's performance after each trial. 
We linearly updated the parameters with a change rate akin to the multi-trial update function from~\cite{Pehlivan2016Minimal}:
\begin{equation}
    \label{eq:maan_adjust_theta}
    \bm\theta_{t+1} = (1 + \bm{\chi}_t)\bm\theta_{t},
\end{equation}
where $t$ is the trial index and $\bm{\chi}_t \in \mathbb{R}^3$ is the change rate vector, which is updated using the two most recent trials:
\begin{equation} \label{eq:related_sc:maan_adjust_change}
\chi_t^i = \chi_{\mathrm{nom}} \frac{\bar{r}_t^i - r_d^i}{r_d^i}\left(\frac{|\bar{r}_t^i - r_d^i|}{|\bar{r}_{t-1}^i - r_d^i|}\right)^{\displaystyle \operatorname{sign}(r_d^i - \bar{r}_t^i)},
\end{equation}
where $r_d^i$ is the $i^\mathrm{th}$ component of the desired error vector $\mathbf{r}_d$ (pre-calculated from an expert demonstration), 
${\bar{r}^i_t}$ is the $i^\mathrm{th}$ component of the average error vector ${\bar{\mathbf{r}}_t}$ during trial $t$, and $\chi_\mathrm{nom}$ is the nominal change rate, which determines the maximum change rate within one update step.

Distance from the end-effector to the closest point along the wire is a suitable performance metric to denote safety in the buzz wire game. This error is conveniently contained in the assistive wrench, $\mathbf{w}_{a}$, such that $\bar{\mathbf{r}}_t$ can be expressed as:
\begin{equation}
\label{eq:r_t}
   \bar{\mathbf{r}}_t = \mathbf{W}^\dagger \overline{\mathbf{w}}_{a, t},
\end{equation}
where $\overline{\mathbf{w}}_{a,t} \in \mathbb{R}^{6}$ is the average wrench from trial $t$ and $\mathbf{W}^\dagger$ is the pseudo-inverse of the coefficient matrix in~\cref{eq:arbitration-mask}.


\subsection{Evaluation Metrics}
\label{sec:exp:metrics}

Performance evaluation for the buzz wire game primarily revolves around two measures: successful task completion time and number of collisions. A trial is deemed successful and collisions are counted when the participant teleoperates the robot's end-effector to the game end (a pre-defined location along the wire) without encountering a fatal failure,~\ie, when collision forces exceed a threshold. Aside from task time and collisions, we also examine a user's \textit{jerk} (the mean squared measure from~\cite{Hogan2009Sensitivity}) in handling the VR controller to capture smoothness or fluency. Moreover, we explore the learning trajectories of users,~\ie, each user's change in ``skill'' metrics relative to their first session. Lastly, we investigate a user's perceptions of the different control modes in terms of \textit{skill}, \textit{success}, and \textit{difficulty}.

Therefore, the hypotheses for this study are as follows:
\begin{itemize}
    \item \textbf{H1:} Subjects will report higher ratings on skill, success, and difficulty for \textbf{sc\_user} than \textbf{teleop} and \textbf{sc}.
    \item \textbf{H2:} There will be faster task success times and less collisions for \textbf{sc\_user} than \textbf{teleop} and \textbf{sc}.
    \item \textbf{H3:} Participants in \textbf{sc\_user} will exhibit better teleoperation performance relative to their `baseline' session when compared against users from \textbf{teleop} and \textbf{sc}.
\end{itemize}

\section{Results}
\label{sec:results}

In the following, we report on our results and test our hypotheses. When evaluating the learning trajectories of participants, we term the first session as the `baseline', the combined second and third sessions as `training', and the last session as `transfer'. Unless otherwise stated, mixed ANOVAs were performed to analyze the effects of control modes (between-subjects) and sessions (within-subjects) on the relevant evaluation metrics. The `baseline' session was left out of the ANOVAs, leaving only the combined `training' sessions to compare against the `transfer' condition.

\subsection{Subjective Results}
\label{sec:results:subj}

\begin{table}[t]
\caption{Subjective results across sessions (1 = ``very low'' or ``failure'', 10 = ``very high'' or ``perfect'')}
\label{tab:session_vis}
\centering
\begin{tabular}{ccccc}
\toprule
Sessions & Mode & Skill $\uparrow$ & Success $\uparrow$ & Difficulty $\downarrow$ \\
\toprule
\multirow{3}{*}{\begin{tabular}[c]{@{}c@{}}1 \\ (baseline)\end{tabular}} & sc & \textbf{7.5 $\pm$ 1.0} & \textbf{7.5 $\pm$ 1.3} & \textbf{4.8 $\pm$ 2.2} \\
& sc\_user & 6.3 $\pm$ 0.5 & 6.0 $\pm$ 1.8 & 6.5 $\pm$ 1.7 \\
& teleop & 6.8 $\pm$ 1.0 & \textbf{7.5 $\pm$ 1.0} & 5.5 $\pm$ 2.4 \\
\midrule
\multirow{3}{*}{\begin{tabular}[c]{@{}c@{}}2 \& 3 \\ (training)\end{tabular}} & sc &  6.1 $\pm$ 3.9 & 6.0 $\pm$ 1.6 & 5.9 $\pm$ 1.8 \\ 
& sc\_user &  7.1 $\pm$ 3.8 & 7.4 $\pm$ 1.4 & 6.1 $\pm$ 1.4 \\
& teleop &  \textbf{7.6 $\pm$ 3.6} & \textbf{7.5 $\pm$ 1.4} & \textbf{5.8 $\pm$ 2.4} \\
\midrule
\multirow{3}{*}{\begin{tabular}[c]{@{}c@{}}4 \\ (transfer)\end{tabular}}& sc & 6.0 $\pm$ 0.8 & 6.8 $\pm$ 1.0 & \textbf{5.0 $\pm$ 1.8} \\
& sc\_user & \textbf{7.3 $\pm$ 0.5} & \textbf{7.0 $\pm$ 0.8} & 5.8 $\pm$ 2.1  \\
& teleop & 6.5 $\pm$ 2.4 & 6.5 $\pm$ 2.4 & 7.0 $\pm$ 3.6  \\
\bottomrule
\end{tabular}
\vspace{-1.2mm}
\end{table}


\begin{table}[t]
    \caption{Subject responses to overall experiment questions (1 = ``strongly disagree'', 5 = ``strongly agree'')}
    \label{tab:general_vis}
    \footnotesize%
    \setlength\tabcolsep{3.5pt} 
    \centering%
    \begin{tabular}{l|ccc}
    \hline
    \hline
    \textbf{Question} & \textbf{Teleop} & \textbf{SC} & \textbf{SC-User}\\
    \hline
    I found the spider chart \textit{helpful} & - & 3.5 $\pm$ 1.0 & \textbf{4.3 $\pm$ 1.0} \\
    \hline
    Most would learn this robot \textit{quickly} & 4.0 $\pm$ 0.8 & 3.8 $\pm$ 0.5 & \textbf{4.5 $\pm$ 0.6} \\
    \hline
    The last session was \textit{more difficult} & 3.3 $\pm$ 1.7 & \textbf{2.5 $\pm$ 0.6} & 3.0 $\pm$ 1.2 \\
    \bottomrule
    \end{tabular}
    \vspace{-5mm}
\end{table}


Subject ratings to post-session questions on \textit{skill} (``How would you rate your skill?''), \textit{success} (``How successful were you?''), and \textit{difficulty} (``How hard did you have to work?'') are shown in~\cref{tab:session_vis}. For the `baseline' session, \textbf{sc} subjects reported better ratings than the other groups, especially in terms of difficulty, with a mean difference of $1.7$ compared to \textbf{sc\_user}. There is a tendency toward more favorable responses for \textbf{teleop} and \textbf{sc\_user} in the `training' sessions, with \textbf{teleop} receiving the best ratings across all categories. Nevertheless, \textbf{sc\_user} has the superior `transfer' according to perceived skill ($7.3\pm0.5$) and success ($7.0\pm0.8$). The \textbf{teleop} users found the new wire testbed more difficult to complete ($7.0\pm3.6$), with a mean difference of $2.0$ from \textbf{sc} and $1.2$ from \textbf{sc\_user}. Despite these patterns, no main or interaction effects were revealed from mixed ANOVAs performed on each question category and so we cannot substantiate \textbf{H1}.

Survey questions on the entire experiment are summarized in~\cref{tab:general_vis}. Indeed, the last session is voted as marginally more difficult by \textbf{teleop} participants. Subjects in the \textbf{sc\_user} group also viewed operating the robot as quicker to learn than \textbf{sc} and even \textbf{teleop}. Interestingly, \textbf{sc\_user} participants considered the spider chart to be more helpful ($4.3\pm1.0$) than the \textbf{sc} chart ($3.5\pm1.0$), which had autonomously computed values and could not be edited. All \textbf{sc\_user} individuals ``found the VR interface easy to edit'' ($5.0$, not in table). 

\subsection{Quantitative Results}
\label{sec:results:quant}

\begin{figure}[t]
    \centering
    \subfigure[Time-to-success per session]{
        \includegraphics[width=0.32\textwidth]{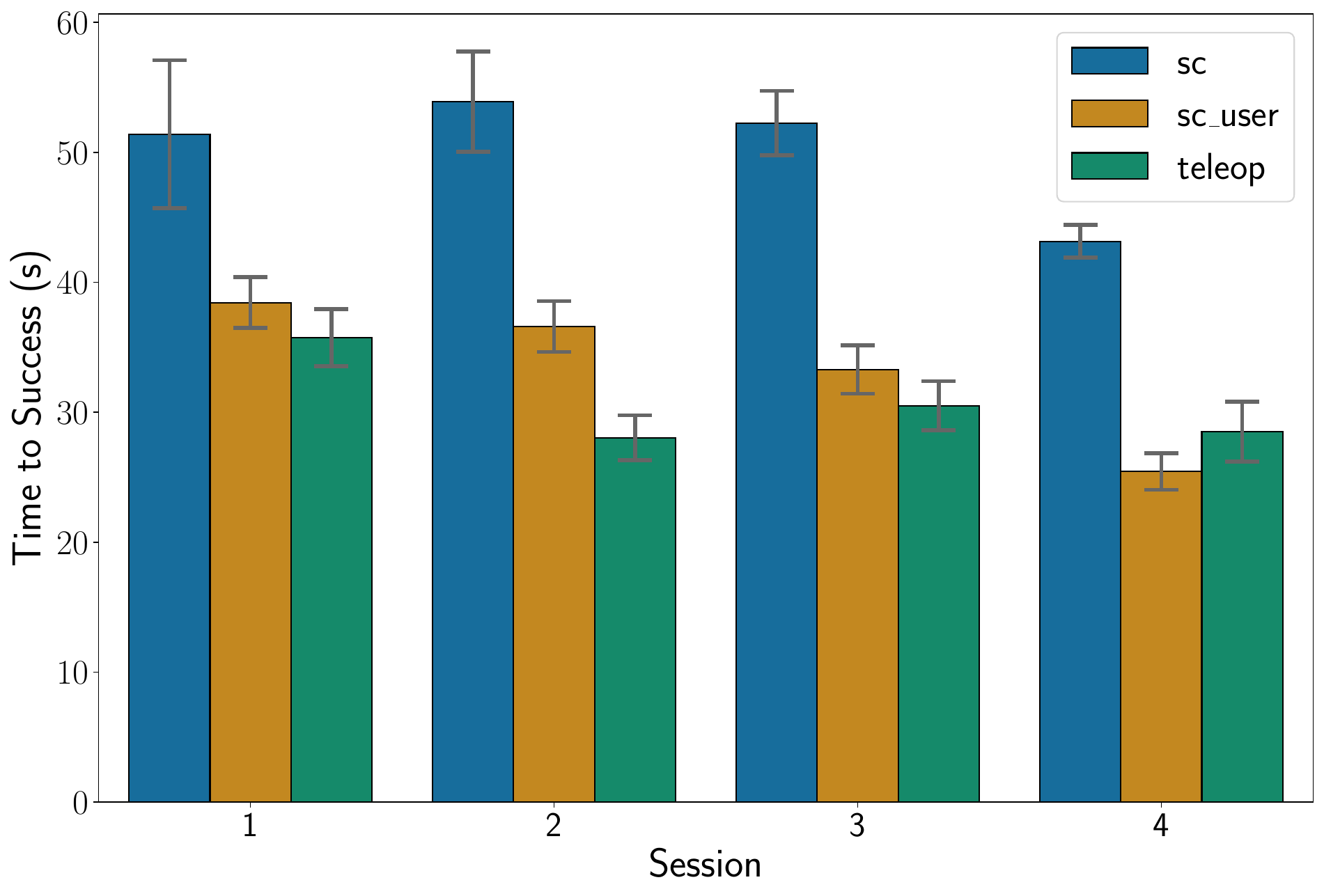}
        \label{fig:tts:session}
    }
    \subfigure[Time-to-success per trial]{
        \includegraphics[width=0.32\textwidth]{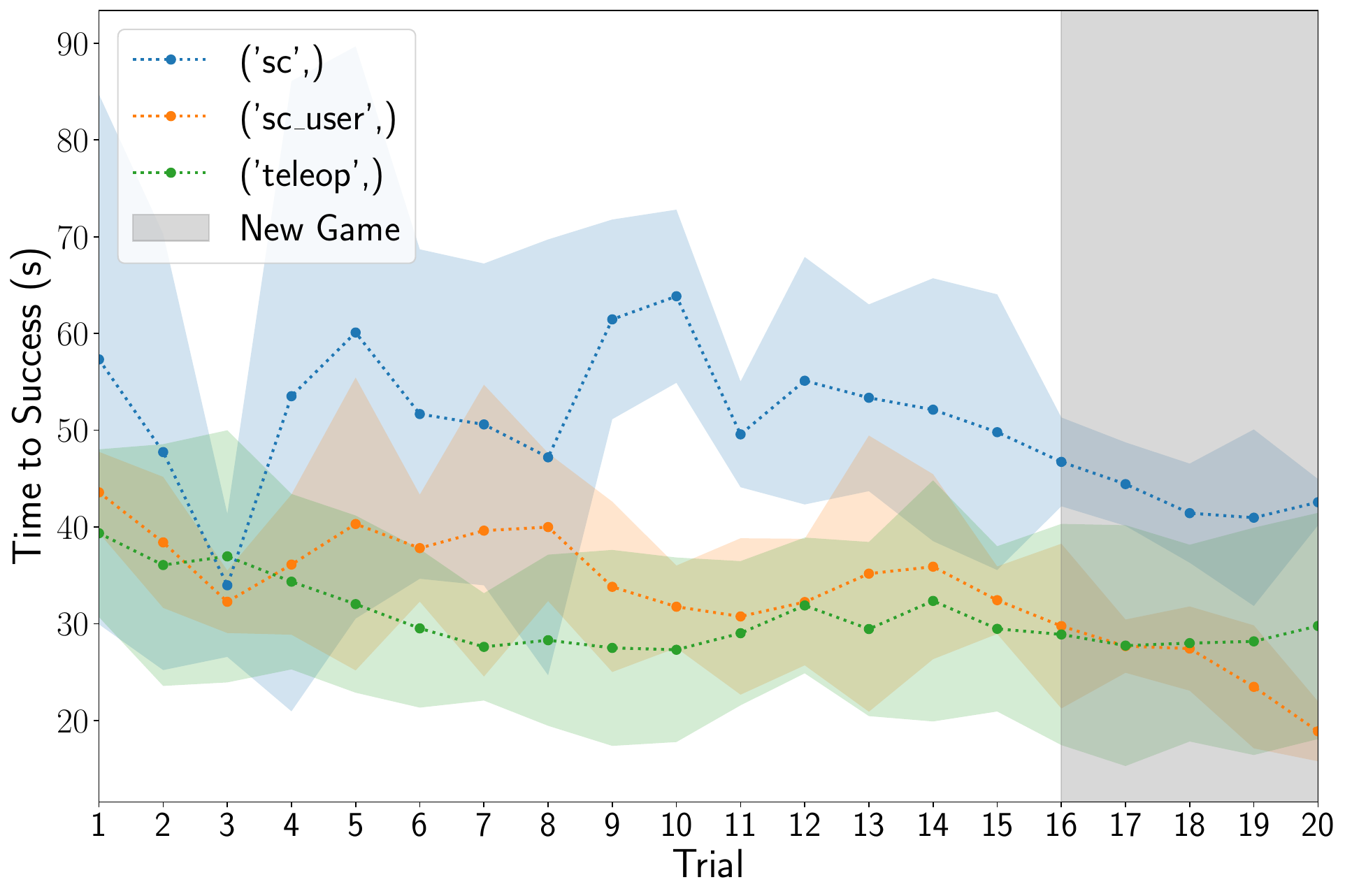}
        \label{fig:tts:trial}
    }

    \caption{Time-to-success demonstrated for participants across (a) sessions and (b) trials. Gray shaded area represents the session with a new test bed.}
    \label{fig:absolute}
    \vspace{-0.8mm}
\end{figure}

\begin{table}[t]
\centering
\caption{Quantitative results across sessions}
\label{tab:quant-overview}
\begin{tabular}{ccccc}
\toprule
Sessions & Mode & Collisions  & Time (s) & No. Fails\\
\toprule
\multirow{3}{*}{\begin{tabular}[c]{@{}c@{}}1 \\ (baseline)\end{tabular}} & sc & \textbf{1.74} $\pm$ \textbf{2.6} & 51.4 $\pm$ 24.8 & 1 \\
& sc\_user & 2.74 $\pm$ 1.9 & 38.4 $\pm$ 8.5 & 1 \\
& teleop & 2.10 $\pm$ 1.8 & \textbf{35.8} $\pm$ \textbf{9.8} & 0 \\
\midrule
\multirow{3}{*}{\begin{tabular}[c]{@{}c@{}}2 \& 3 \\ (training)\end{tabular}} & sc & 1.60 $\pm$ 2.0 & 53.1 $\pm$ 12.8 & 6 \\
& sc\_user & \textbf{1.50} $\pm$ \textbf{1.3} & 35.0 $\pm$ 8.5 & 0 \\
& teleop & 1.54 $\pm$ 1.4 & \textbf{29.3} $\pm$ \textbf{8.0} & 1 \\
\midrule
\multirow{3}{*}{\begin{tabular}[c]{@{}c@{}}4 \\ (transfer)\end{tabular}}& sc & 2.26 $\pm$ 1.9 & 
43.2 $\pm$ 5.4 & 1 \\
& sc\_user & \textbf{1.45} $\pm$ \textbf{1.1}  & \textbf{25.5} $\pm$ \textbf{6.3} & 0 \\
& teleop & 1.70 $\pm$ 1.4  & 28.5 $\pm$ 10.3 & 0 \\
\bottomrule
\end{tabular}
\vspace{-3mm}
\end{table}

\cref{tab:quant-overview} summarizes the quantitative results of the study. In the first `baseline' session, the \textbf{sc} group had the least number of collisions ($1.74\pm2.6$), whereas \textbf{teleop} had the fastest time-to-success ($35.8\text{s}\pm9.8\text{s}$). 
On the other hand, \textbf{sc\_user} subjects exhibited notable progress in `training' sessions, as there was a major reduction in average number of collisions, yielding the lowest collision count ($1.50\pm1.3$). The \textbf{teleop} group continued to maintain the fastest success times ($29.3\text{s}\pm8.0\text{s}$). 
Although in the `transfer' session, \textbf{sc\_user} participants attained the swiftest completion times ($25.5\text{s}\pm6.3\text{s}$) and the fewest collisions ($1.45\pm 1.1$). 

Average time-to-success is presented per session and trial in~\cref{fig:tts:session,fig:tts:trial}, respectively.
A mixed ANOVA run on completion times found no interaction effect between modes and sessions ($F(2,9) \, {=} \, 1.17, \, p \, {=} \, 0.35$). Though a main effect was found for both mode ($p \, {<} \, 0.01$) and session ($p \, {<} \, 0.05$). Post-hoc pairwise comparisons validated that the \textbf{sc} times were slower than those for both \textbf{teleop} ($p \, {<} \, 0.01$) and \textbf{sc\_user} ($p \, {<} \, 0.001$). A paired post-hoc test also suggests that completion times were faster in the final session ($p \, {<} \, 0.05$), hinting at the `transfer' task being a quicker game to navigate. These results only partially support \textbf{H2}, as \textbf{teleop} and \textbf{sc\_user} obtained statistically similar times ($p \, {=} \, 0.81$). We still highlight the trend in \textbf{sc\_user} operators displaying faster task times compared to the \textbf{teleop} group when transitioning to the new testbed. Smaller standard deviations also bolster this trend, as shown in~\cref{fig:tts:trial}.


\begin{figure}[t]
    \centering
    \subfigure[Collisions vs baseline per session]{
        \includegraphics[width=0.32\textwidth]{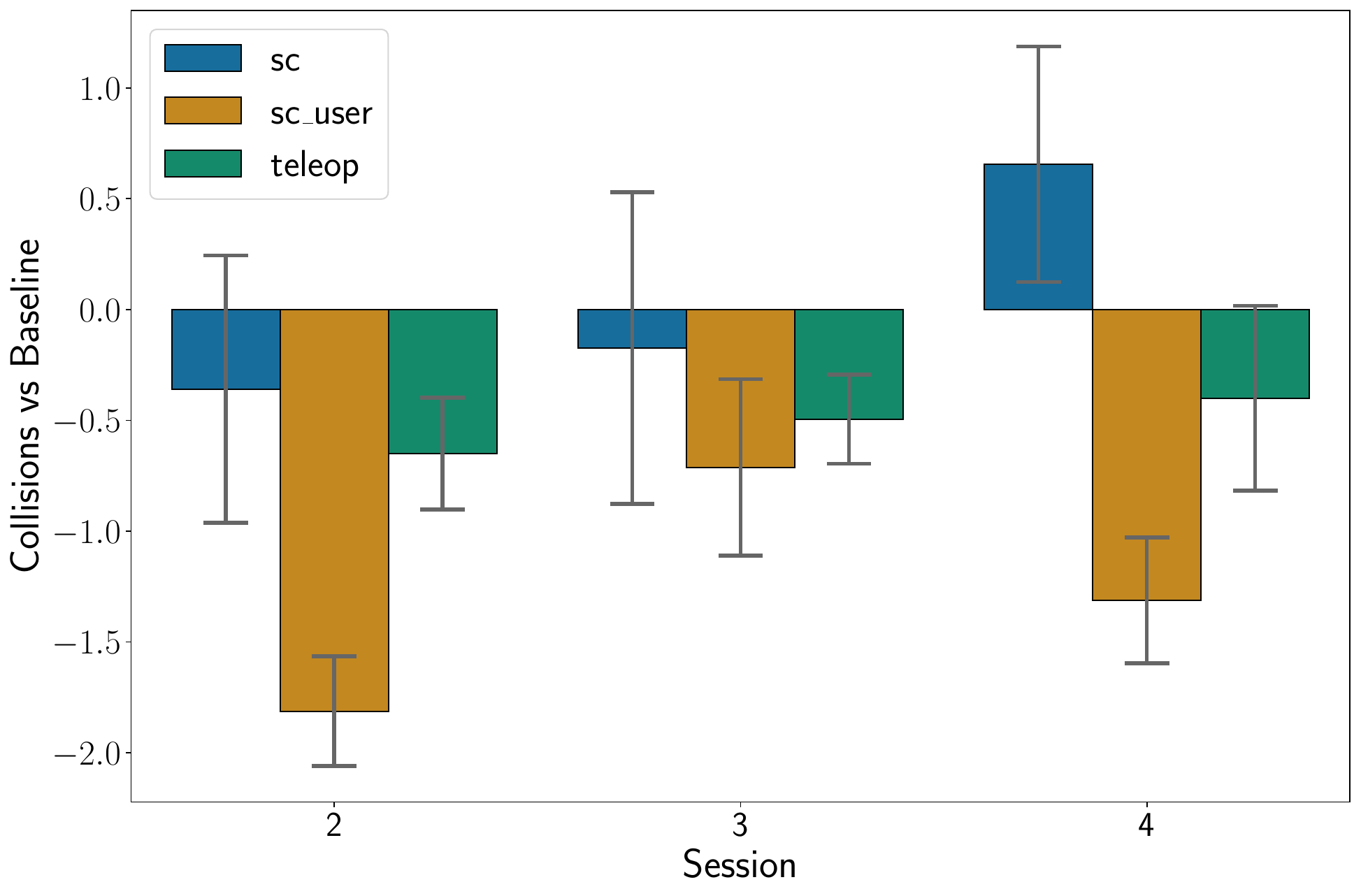}
        \label{fig:coll_baseline}
    }
    \subfigure[Control interface jerk vs baseline per session]{
        \includegraphics[width=0.32\textwidth]{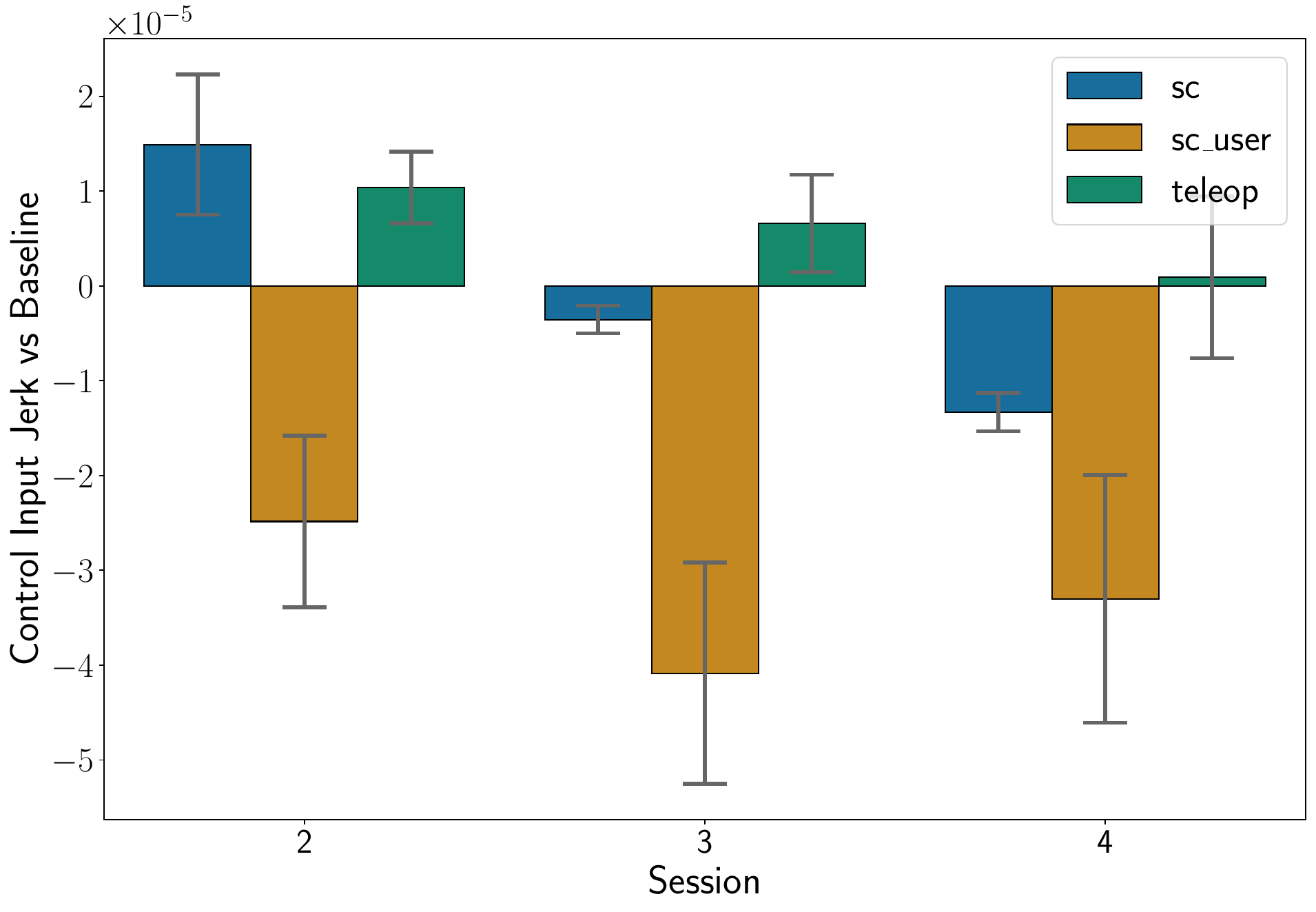}
        \label{fig:rel_jerk}
    }
    \caption{Differences in collision occurrences (a) and control interface jerk (b) \textit{relative} to the subject's average baseline performance in the first session.}
    \label{fig:relative}
    \vspace{-5.3mm}
\end{figure}

Comparing a subject's teleoperation performance between their `baseline' session and subsequent sessions is informative of their learning trajectory. \cref{fig:coll_baseline,fig:rel_jerk} illustrate these trajectories for collision frequency and controller jerk, respectively, which we use to quantify operator ``skill''. A mixed ANOVA for collisions found no interaction or main effects in modes and session. Yet a one-way ANOVA to observe if the control mode in the `transfer' session influenced collisions relative to the subject's `baseline' revealed a significant effect ($F(2,56) \, {=} \, 5.47, \, p \, {<} \, 0.01$). Post-hoc comparisons showed that the only significant effect was \textbf{sc\_user} incurring fewer relative collisions than \textbf{sc} ($p \, {<} \, 0.01$). Similarly, a mixed ANOVA found no interactions or main differences on control interface jerk, but a one-way ANOVA found a significant effect of mode in the final session ($F(2,57) \, {=} \, 3.52, \, p \, {<} \, 0.05$). A post-hoc test indicated that \textbf{sc\_user} issued less jerky commands than \textbf{teleop} ($p \, {<} \, 0.05$). Overall, these findings offer partial support for \textbf{H3}.

For the analyses discussed above, we did not observe any significant effects of prior robotics experience when isolating control strategies. Even one-way ANOVAs on the `baseline' session (where significance would be most likely) found no differences,~\eg, task times for subjects in \textbf{teleop} with robotics background were not faster than those without ($p \, {=} \, 0.066$). Nonetheless, we cannot draw conclusions about these results given the small number of participants.

\section{Discussion}
\label{sec:discuss}

A few notable findings emerged from the buzz wire experiment. First, while the subjective and \textit{absolute} results were comparable for \textbf{sc\_user} and \textbf{teleop} during  `training' sessions, \textbf{sc\_user} operators outperformed the other control modes in measures of \textit{relative} improvement. For example, \textbf{sc\_user} subjects had consistently less collisions and produced less jerky inputs than their `baseline' session, which were statistically significant differences over \textbf{teleop} and \textbf{sc} in the `transfer' condition. Indeed, \textbf{sc\_user} obtained markedly better results across all subjective and quantitative (absolute and relative) metrics in the new game. Altogether, these results reinforce the benefits of offering users a means of customizing shared control, particularly when there are variations in the task.

\begin{figure}[t]
    \centering
    \includegraphics[width=0.88\columnwidth]{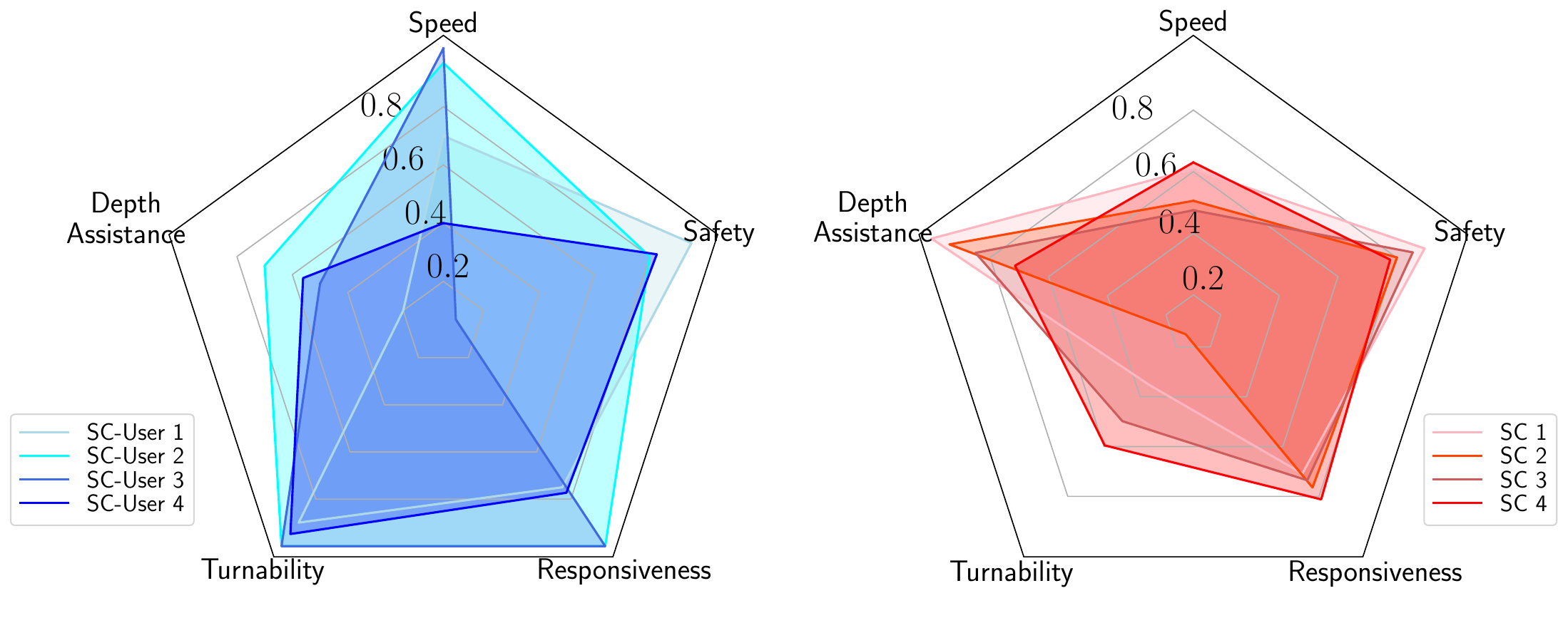}
    \caption{User-customized vs heuristics-based arbitration parameters at the conclusion of the experiment.}
    \label{fig:radar_results}
    \vspace{-4.9mm}
\end{figure}

Another observation is the inferior performance of the \textbf{sc} policy. Comparing the distributions of arbitration parameters between $\textbf{sc\_user}$ and $\textbf{sc}$ reveals higher variance in the $\textbf{sc\_user}$ group's final $\bm{\theta}$ selection, as depicted in~\cref{fig:radar_results}. The invariance in $\textbf{sc}$ is due to the performance heuristic $\bar{\mathbf{r}}_t$ from~\cref{eq:r_t} being solely dependent on the distance between the end-effector and wire. An alternative update routine could prioritize factors that  $\textbf{sc\_user}$ participants gravitated toward, such as \textit{turnability} and \textit{responsiveness}. However, any heuristics-based method for automatically adapting $\bm{\theta}$ would still converge to a similar parameter distribution across users, without accounting for individual preferences. Shared control policies learned from large demonstration datasets would also suffer from the lack of personalization to unseen users. 


Despite our promising preliminary findings, there are areas for improvement. One important limitation is the broader applicability of our results due to the constrained task and environment. Future research should explore more complex telemanipulation tasks,~\eg, those involving humanoids~\cite{darvish2023teleoperation}, and include a larger, more diverse participant pool with varied backgrounds and skill levels in robot teleoperation. Another imperative requirement going forward will be to ensure that users have a clear understanding of the implications in visualization signals, $\mathcal{V}$, and interface actions, $v_h$. Regardless of our best efforts to make $\mathcal{V}$ and $v_h$ intuitive, there was one occasion where an \textbf{sc\_user} subject misunderstood the meaning of a parameter, leading to a deterioration in performance. Simplifying the selection of arbitration factors may alleviate this issue, especially since certain factors,~\eg, \textit{turnability}, were redundant as all \textbf{sc\_user} subjects converged on a similar setting (see~\cref{fig:radar_results}).

Finally, we suggest ways to extend our user-customizable formulation to make it more widely applicable. Instead of using various task-specific factors, like \textit{speed}, a single scalar $\alpha \in [0,1]$ for ``autonomous assistance'' could be easily tuned in VR,~\eg, with a graphical slider. Modern frameworks for shared control, such as those based on reinforcement learning~\cite{Reddy2018Shared}, could then let users calibrate $\alpha$ \textit{online}, rather than relying on a fixed or learned value that they cannot see or refine. Alternatively, latent actions learned from expert demonstrations could be made user-adjustable~\cite{Cui2023No,Jeon2020Shared}. The main challenges here would be in communicating the latent dimensions through $\mathcal{V}$ and exposing user interface actions to modify their values, as these latent actions may not represent a human-interpretable factor or be easy to edit via $v_h$. We defer investigation into these ideas for future work.





\section{Conclusions}

In this paper, we introduced a mathematical shared control framework for users to customize the arbitration process. We presented an instantiation of this framework for a teleoperation task in $\mathrm{SE}(3)$ and conducted a longitudinal study spanning two weeks per subject to evaluate the proposed method. Our user-customizable shared control method enhanced teleoperation outcomes, reducing collision frequency and input jerk, while also accommodating the baseline skill levels of different users. Furthermore, operators transferred to generalizations in the teleoperation task better when they could edit the inner workings of the shared control. We believe these results hold promise for long-term adaptability to diverse variations in users and tasks. Future research will extend the scale of the user study, explore a wider array of tasks, and consider more flexible representations of the arbitration parameters for communication and user editing.



\bibliographystyle{IEEEtran}
\bibliography{IEEEabrv,references}

\end{document}